\documentclass[10pt,twocolumn,letterpaper]{article}

\usepackage{iccv}
\usepackage{times}
\usepackage{epsfig}
\usepackage{graphicx}
\usepackage{amsmath}
\usepackage{amssymb}
\usepackage[accsupp]{axessibility}
\usepackage{caption}
\usepackage{makecell}

\usepackage[breaklinks=true,bookmarks=false]{hyperref}
\usepackage{url}
\iccvfinalcopy 

\def\iccvPaperID{****} 

\ificcvfinal\fi

\begin{document}

\title{Oriented R-CNN for Object Detection}

\author{Xingxing Xie \hspace{3mm} Gong Cheng* \hspace{3mm} Jiabao Wang \hspace{3mm} Xiwen Yao \hspace{3mm} Junwei Han\\
School of Automation, Northwestern Ploytechnical University, Xi'an, China\\
{\tt\small \{xiexing,jbwang\}@mail.nwpu.edu.cn \hspace{0.5mm} \{gcheng,yaoxiwen,jhan\}@nwpu.edu.cn}
\vspace{-2.5mm}
}

\twocolumn[{%
	\renewcommand\twocolumn[1][]{#1}%
	\maketitle
	\begin{center}
		\centering
		\includegraphics[width=0.95\textwidth]{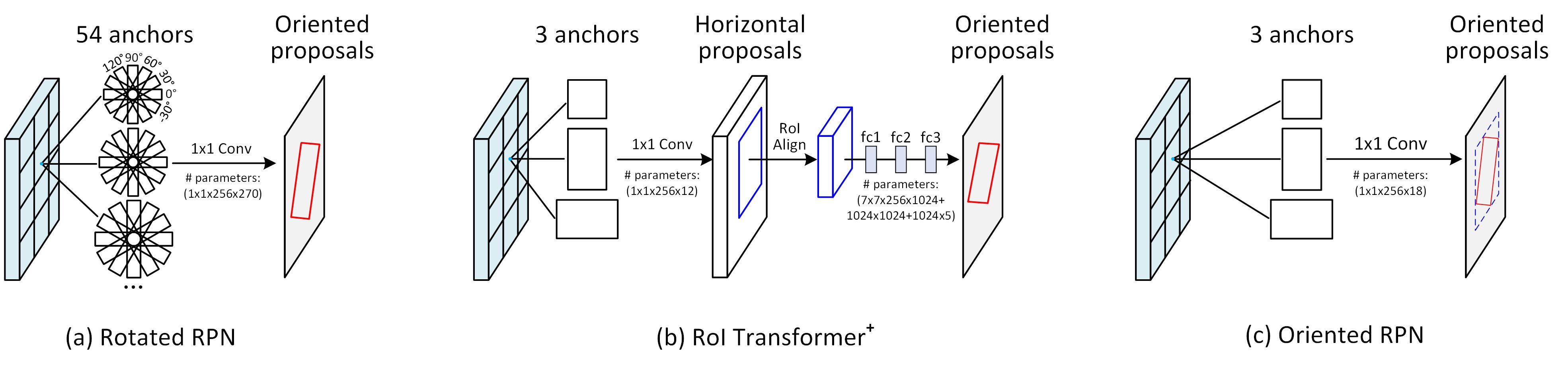}
		\vspace*{-2mm}
		\captionof{figure}{Comparisons of different schemes for generating oriented proposals. (a) Rotated RPN densely places rotated anchors with different scales, ratios, and angles. (b) RoI Transformer$^\textbf{+}$ learns oriented proposal from horizontal RoI. It involvs RPN, RoI Alignment, and regression. (c) Our proposed oriented RPN generates high-quality proposals in a nearly cost-free manner. The number of parameters of oriented RPN is about 1/3000 of RoI Transformer$^\textbf{+}$ and 1/15 of rotated RPN.}
		\label{Fig1}
		\vspace{0.5\baselineskip}
	\end{center}%
}]

\maketitle
\ificcvfinal\thispagestyle{empty}\fi

\renewcommand{\thefootnote}{}
\begin{abstract}
   \footnotetext{*Corresponding author. \hspace{0.4mm}  $^\textbf{+}$ denotes the official version of RoI Transformer implemented on AerialDetection (the same below).}Current state-of-the-art two-stage detectors generate oriented proposals through time-consuming schemes. This diminishes the detectors' speed, thereby becoming the computational bottleneck in advanced oriented object detection systems. This work proposes an effective and simple oriented object detection framework, termed Oriented R-CNN, which is a general two-stage oriented detector with promising accuracy and efficiency. To be specific, in the first stage, we propose an oriented Region Proposal Network (oriented RPN) that directly generates high-quality oriented proposals in a nearly cost-free manner. The second stage is oriented R-CNN head for refining oriented Regions of Interest (oriented RoIs) and recognizing them. Without tricks, oriented R-CNN with ResNet50 achieves state-of-the-art detection accuracy on two commonly-used datasets for oriented object detection including DOTA (75.87\% mAP) and HRSC2016 (96.50\% mAP), while having a speed of 15.1 FPS with the image size of 1024$\times$1024 on a single RTX 2080Ti. We hope our work could inspire rethinking the design of oriented detectors and serve as a baseline for oriented object detection. Code is available at \url{https://github.com/jbwang1997/OBBDetection}.
\end{abstract}

\section{Introduction}
Most existing state-of-the-art oriented object detection methods~\cite{ding2019, xu2020, yang2019} depend on proposal-driven frameworks, like Fast/Faster R-CNN~\cite{girshick2014,girshick2015,ren2016}. They involve two key steps: (i) generating oriented proposals and (ii) refining proposals and classifying them into different categories. Nevertheless, the step of producing oriented proposals is computationally expensive. 

One of the early methods of generating oriented proposals is rotated Region Proposal Network (rotated RPN for short)~\cite{ma2018}, which places 54 anchors with different angles, scales and aspect rations (3 scales$\times$3 ratios$\times$6 angles) on each location, as shown in Figure \ref{Fig1}(a). The introduction of rotated anchors improves the recall and demonstrates good performance when the oriented objects distribute sparsely. However, abundant anchors cause massive computation and memory footprint. To address this issue, RoI Transformer~\cite{ding2019} learns oriented proposals from horizontal RoIs by complex process, which involves RPN, RoI Alignment and regression (see Figure \ref{Fig1}(b)). The RoI Transformer scheme provides promising oriented proposals and drastically reduces the number of rotated anchors, but also brings about expensive computation cost. Now, how to design an elegant and efficient solution to generate oriented proposals is the key to breaking the computational bottleneck in state-of-the-art oriented detectors.

To push the envelope further: we investigate why the efficiency of region proposal-based oriented detectors has trailed thus far. Our observation is that the main obstacle impeding the speed of proposal-driven detectors is from the stage of proposal generation. A natural and intuitive question to ask is: can we design a general and simple oriented region proposal network (oriented RPN for short) to generate high-quality oriented proposals directly? Motivated by this question, this paper presents a simple two-stage oriented object detection framework, called oriented R-CNN, which obtains state-of-the-art detection accuracy, while keeping competitive efficiency in comparison with one-stage oriented detectors.

To be specific, in the first stage of oriented R-CNN, a conceptually simple oriented RPN is presented (see Figure \ref{Fig1}(c)). Our oriented RPN is a kind of light-weight fully convolutional network, which has extremely fewer parameters than rotated RPN and RoI transformer$^\textbf{+}$, thus reducing the risk of overfitting. We realize it by changing the number of output parameters of RPN regression branch from four to six. There is no such thing as a free lunch. The design of oriented RPN benefits from our proposed representation scheme of oriented objects, named midpoint offset representation. For arbitrary-oriented objects in images, we utilize six parameters to represent them. The midpoint offset representation inherits horizontal regression mechanism, as well as provides bounded constraints for predicting oriented proposals. The second stage of oriented R-CNN is oriented R-CNN detection head: extracting the features of each oriented proposal by rotated RoI alignment and performing classification and regression.

Without bells and whistles, we evaluate our oriented R-CNN on two popular benchmarks for oriented object detection, namely DOTA and HRSC2016. Our method with ResNet-50-FPN surpasses the accuracy of all existing state-of-the-art detectors, achieving 75.87\% mAP on the DOTA dataset and 96.50\% mAP on the HRSC2016 dataset, while running at 15.1 FPS with the image size of 1024$\times$1024 on a single RTX 2080Ti. Thus, the oriented R-CNN is a practical object detection framework in terms of both accuracy and efficiency. We hope our method will inspire rethinking the design of oriented object detectors and oriented object regression scheme, as well as serve as a solid baseline for oriented object detection. For future research, our code is available at \url{https://github.com/jbwang1997/OBBDetection}.

\section{Related Work}
In the past decade, remarkable progresses~\cite{liu2020deep,tian2019fcos,cai2018cascade,yang2019reppoints,redmon2016you,liu2016ssd,zhang2020bridging,he2017mask,li2020object,pang2019libra} have been made in the field of object detection. As an extended branch of object detection, oriented object detection~\cite{ding2019,xu2020,ma2018,han2021,cheng2016learning,pan2020dynamic} has received extensive attention driven by its wide applications.

General object detection methods (e.g., Faster R-CNN) relying on horizontal bounding boxes could not tightly locate oriented objects in images, because a horizontal bounding box may contain more background regions or multiple objects. This results in the inconsistency between the final classification confidence and localization accuracy. To address this issue, much attention has been devoted. For instance, Xia et al.~\cite{xia2018dota} built a large-scale object detection benchmark with oriented annotations, named DOTA. Many existing oriented object detectors~\cite{ding2019,xu2020,yang2019,ma2018,liu2016ship,8803521} are mainly based on typical proposal-based object detection frameworks. A nature solution to oriented object is to set rotated anchors~\cite{ma2018,liu2016ship}, such as rotated RPN~\cite{ma2018}, in which the anchors with different angles, scales and aspect ratios are placed on each location. These densely rotated anchors lead to extensive computations and memory footprint. 

To decrease the large number of rotated anchors and reduce the mismatch between features and objects, Ding et.al~\cite{ding2019} proposed the RoI transformer that learns rotated RoIs from horizontal RoIs produced by RPN. This manner greatly boosts the detection accuracy of oriented objects. However, it makes the network heavy and complex because it involves fully-connected layers and RoI alignment operation during the learning of rotated RoIs. To address the challenges of small, cluttered, and rotated object detection, Yang et.al~\cite{yang2019} built an oriented object detection method on the generic object detection framework of Faster R-CNN. Xu et.al~\cite{xu2020} proposed a new oriented object representation, termed gliding vertexes. It achieves oriented object detection by learning four vertex gliding offsets on the regression branch of Faster R-CNN head. However, these two methods both adopt horizontal RoIs to perform classification and oriented bounding box regression. They still suffer from severe misalignment between objects and features.
\begin{figure*}
	\begin{center}
		\includegraphics[width=0.85\linewidth]{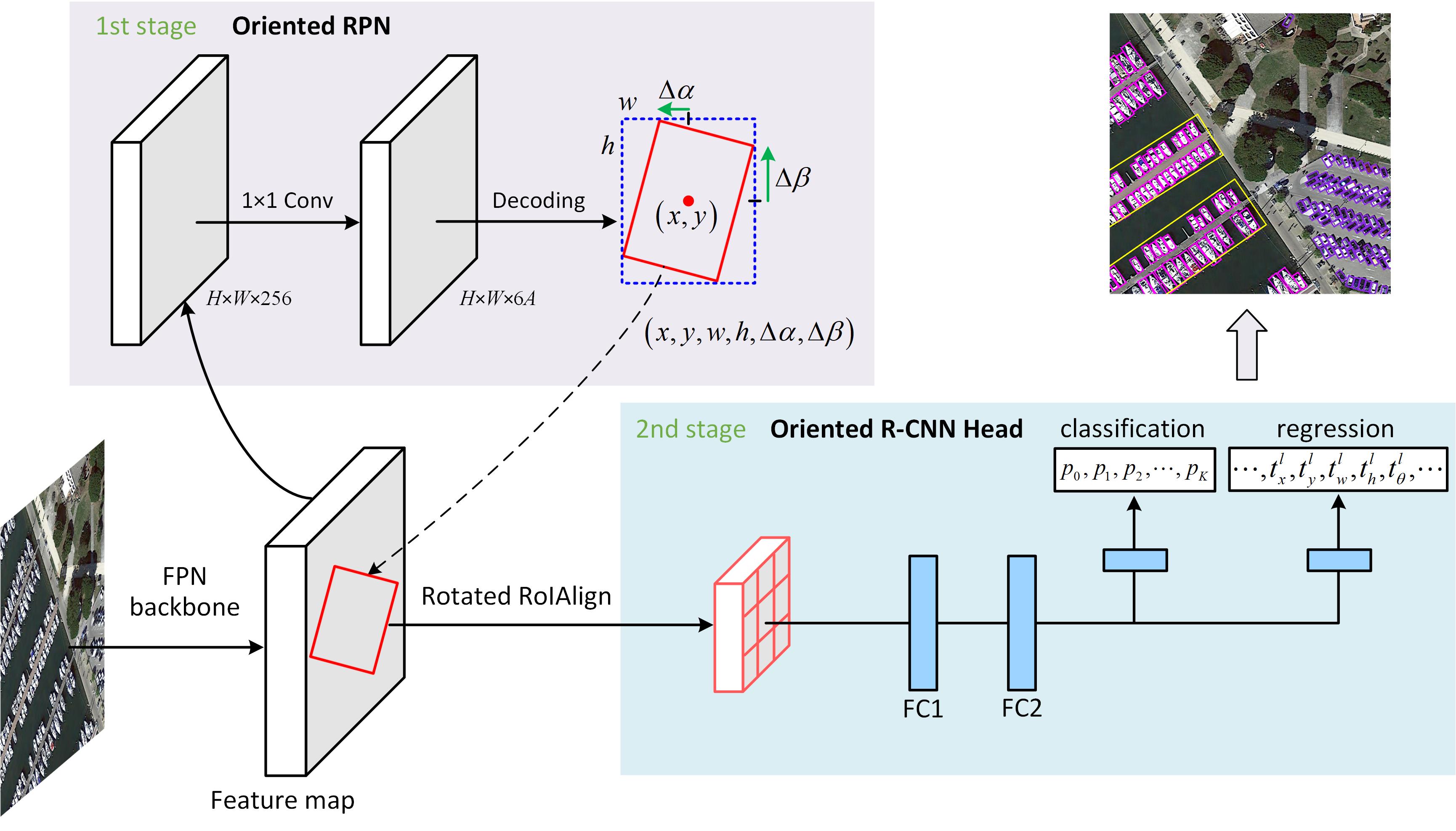}\\
	\end{center}
    \vspace*{-5mm}
	\caption{Overall framework of oriented R-CNN, which is a two-stage detector built on FPN. The first stage generates oriented proposals by oriented RPN and the second stage is oriented R-CNN head to classify proposals and refine their spatial locations. For clear illustration, we do not show the FPN as well as the classification branch in oriented RPN.}\label{Fig2}
	\vspace*{-3mm}
\end{figure*}

In addition, some works~\cite{han2021,pan2020dynamic,yang2019r3det,ming2020dynamic,yi2021oriented,hou2020cascade,yang2020arbitrary,zhou2017east,he2018end} have explored one-stage or anchor-free oriented object detection frameworks: outputting object classes and oriented bounding boxes without region proposal generation and RoI alignment operation. For example, Yang et.al~\cite{yang2019r3det} proposed a refined one-stage oriented object detector, which involves two key improvements, including feature refinement and progressive regression, to address the problem of feature misalignment. Ming et.al~\cite{ming2020dynamic} designed a new label assignment strategy for one-stage oriented object detection based on RetinaNet~\cite{lin2017focal}. It assigns the positive or negative anchors dynamically through a new matching strategy. Han et.al~\cite{han2021} proposed a single-shot alignment network (S$^{2}$ANet) for oriented object detection. S$^{2}$ANet aims at alleviating the inconsistency between the classification score and location accuracy via deep feature alignment. Pan et.al~\cite{pan2020dynamic} devised a dynamic refinement network (DRN) for oriented object detection based on the anchor-free object detection method CenterNet~\cite{zhou2019objects}.

In contrast with the above methods, our work falls within proposal-based oriented object detection methods. It focuses on designing a high-efficiency oriented RPN to break the computational bottleneck of generating oriented proposals.

\section{Oriented R-CNN}
Our proposed object detection method, called oriented R-CNN, consists of an oriented RPN and an oriented R-CNN head (see Figure \ref{Fig2}). It is a two-stage detector, where the first stage generates high-quality oriented proposals in a nearly cost-free manner and the second stage is oriented R-CNN head for proposal classification and regression. Our FPN backbone follows~\cite{lin2017feature}, which produces five levels of features $\left\{P_{2}, P_{3}, P_{4}, P_{5}, P_{6}\right\}$. For simplicity, we do not show the FPN architecture as well as the classification branch in oriented RPN. Next, we describe the oriented RPN and oriented R-CNN head in detail.

\subsection{Oriented RPN}


Given an input image of any size, oriented RPN outputs a sparse set of oriented proposals. The entire process can be modeled by light-weight fully-convolutional networks. 

Specifically, it takes five levels of features $\left\{P_{2}, P_{3}, P_{4}, P_{5}, P_{6}\right\}$  of FPN as input and attaches a head of the same design (a 3$\times$3 convolutional layer and two sibling 1$\times$1 convolutional layers) to each level of features. We assign three horizontal anchors with three aspect rations \{1:2, 1:1, 2:1\} to each spatial location in all levels of features. The anchors have the pixel areas of $\left\{32^2, 64^2, 128^2, 256^2, 512^2\right\}$ on $\left\{P_{2}, P_{3}, P_{4}, P_{5}, P_{6}\right\}$, respectively. Each anchor $\boldsymbol{a}$ is denoted by a 4-dimensional vector $\boldsymbol{a}=\left(a_{x}, a_{y}, a_{w}, a_{h}\right)$, where $\left(a_{x}, a_{y}\right)$ is the center coordinate of the anchor, $a_{w}$ and $a_{h}$ represent the width and height of the anchor.
One of the two sibling 1$\times$1 convolutional layers is regression branch: outputting the offset $\boldsymbol{\delta}=\left(\delta_{x}, \delta_{y}, \delta_{w}, \delta_{h}, \delta_{\alpha}, \delta_{\beta}\right)$ of the proposals relative to the anchors. At each location of feature map, we generate $A$ proposals ($A$ is the number of anchors at each location, and it equals to 3 in this work), thus the regression branch has 6$A$ outputs. By decoding the regression outputs, we can obtain the oriented proposal. The process of decoding is described as follows:
\begin{equation}
\label{equ1}
\left\{\begin{array}{l}
\Delta \alpha=\delta_{\alpha} \cdot w, \quad \Delta \beta=\delta_{\beta} \cdot h \\
w=a_{w} \cdot e^{\delta_{w}}, \quad h=a_{h} \cdot e^{\delta_{h}} \\
x=\delta_{x} \cdot a_{w}+a_{x}, \quad y=\delta_{y} \cdot a_{h}+a_{y}
\end{array}\right.
\end{equation}
where $(x, y)$ is the center coordinate of the predicted proposal, $w$ and $h$ are the width and height of the external rectangle box of the predicted oriented proposal. $\Delta \alpha$ and $\Delta \beta$ are the offsets relative to the midpoints of the top and right sides of the external rectangle. Finally, we produce oriented proposals according to $(x, y, w, h, \Delta \alpha, \Delta \beta)$.

The other sibling convolutional layer estimates the objectness score for each oriented proposal. For clear illustration, we omit the scoring branch in Figure 2. The oriented RPN is acctually a natural and intuitive idea, but its key lies in the representation of oriented objects. Under this circumstance, we design a new and simple representation scheme of oriented objects, called midpoint offset representation.
\vspace*{-2.5mm}

\subsubsection{Midpoint Offset Representation}
\begin{figure}
	\begin{center}
		\includegraphics[width=0.90\linewidth]{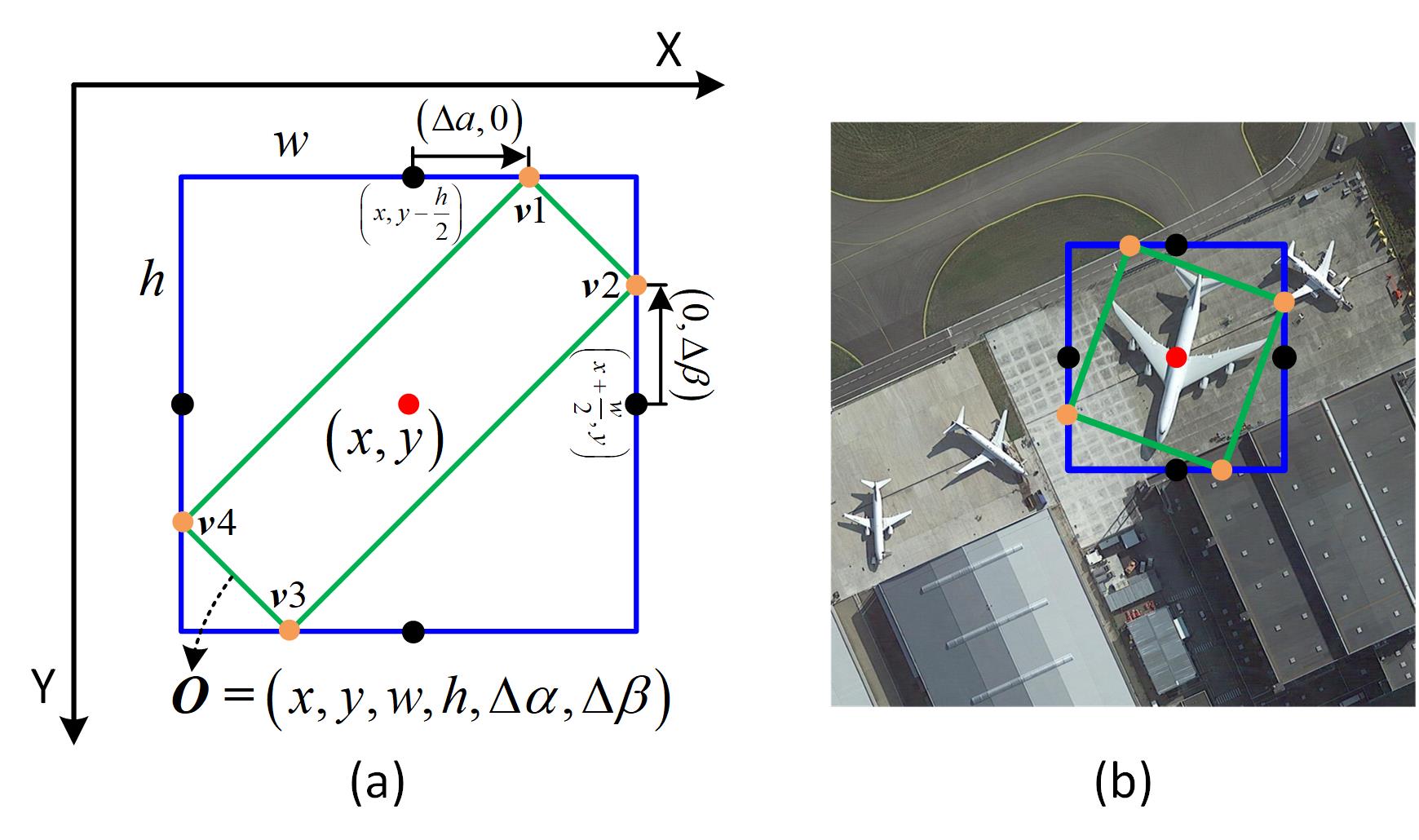}\\
	\end{center}
    \vspace*{-5mm}
	\caption{Illustration of midpoint offset representation. (a) The schematic diagram of midpoint offset representation. (b) An example of midpoint offset representation.}\label{Fig3}
	\vspace{-4mm}
\end{figure}

We propose a novel representation scheme of oriented objects, named midpoint offset representation, as shown in Figure \ref{Fig3}. The black dots are the midpoints of each side of the horizontal box, which is the external rectangle of the oriented bounding box \textit{\textbf{O}}. The orange dots stand for the vertexes of the oriented bounding box \textit{\textbf{O}}.

Specifically, we use an oriented bounding box \textit{\textbf{O}} with six parameters $\textit{\textbf{O}}=(x, y, w, h, \Delta \alpha, \Delta \beta)$  to represent an object computed by Equation (\ref{equ1}). Through the six parameters, we can obtain the coordinate set $\mathbf{v}=(\boldsymbol{v} 1, \boldsymbol{v} 2, \boldsymbol{v} 3, \boldsymbol{v} 4)$ of four vertexes for each proposal. Here, $\Delta \alpha$ is the offset of $\boldsymbol{v}1$ with respect to the midpoint $(x, y-h / 2)$ of the top side of the horizontal box. According to the symmetry, $-\Delta \alpha$ represents the offset of $\boldsymbol{v}3$  with respect to the bottom midpoint $(x, y+h / 2)$. $\Delta \beta$ stands for the offset of $\boldsymbol{v}2$  with respect to the right midpoint $(x+w / 2, y)$, and $-\Delta \beta$ is the offset of $\boldsymbol{v}4$ with respect to the left midpoint $(x-w / 2, y)$. Thus, the coordinates of four vertexes can be expressed as follows.
\begin{equation}
\left\{\begin{array}{l}
\boldsymbol{v} 1=(x, y-h / 2)+(\Delta \alpha, 0) \\
\boldsymbol{v} 2=(x+w / 2, y)+(0, \Delta \beta) \\
\boldsymbol{v} 3=(x, y+h / 2)+(-\Delta \alpha, 0) \\
\boldsymbol{v} 4=(x-w / 2, y)+(0,-\Delta \beta)
\end{array}\right.
\end{equation}

With the representation manner, we implement the regression for each oriented proposal through predicting the parameters $(x, y, w, h)$ for its external rectangle and inferring the parameters $(\Delta \alpha, \Delta \beta)$ for its midpoint offset.
\vspace*{-2.5mm}
\subsubsection{Loss Function}
To train oriented RPN, the positive and negative samples are defined as follows. First, we assign a binary label $p^{*} \in\{0,1\}$ to each anchor. Here, 0 and 1 mean that the anchor belongs to positive or negative sample. To be specific, we consider an anchor as positive sample under one of the two conditions: (i) an anchor having an Intersection-over-Union (IoU) overlap higher than 0.7 with any ground-truth box, (ii) an anchor having the highest IoU overlap with a ground-truth box and the IoU is higher than 0.3. The anchors are labeled as negative samples when their IoUs are lower than 0.3 with ground-truth box. The anchors that are neither positive nor negative are considered as invalid samples, which are ingnored in the training process. It is worth noting that the above-mentioned ground-truth boxes refer to the external rectangles of oriented bounding boxes.
\begin{figure}
	\begin{center}
		\includegraphics[width=0.90\linewidth]{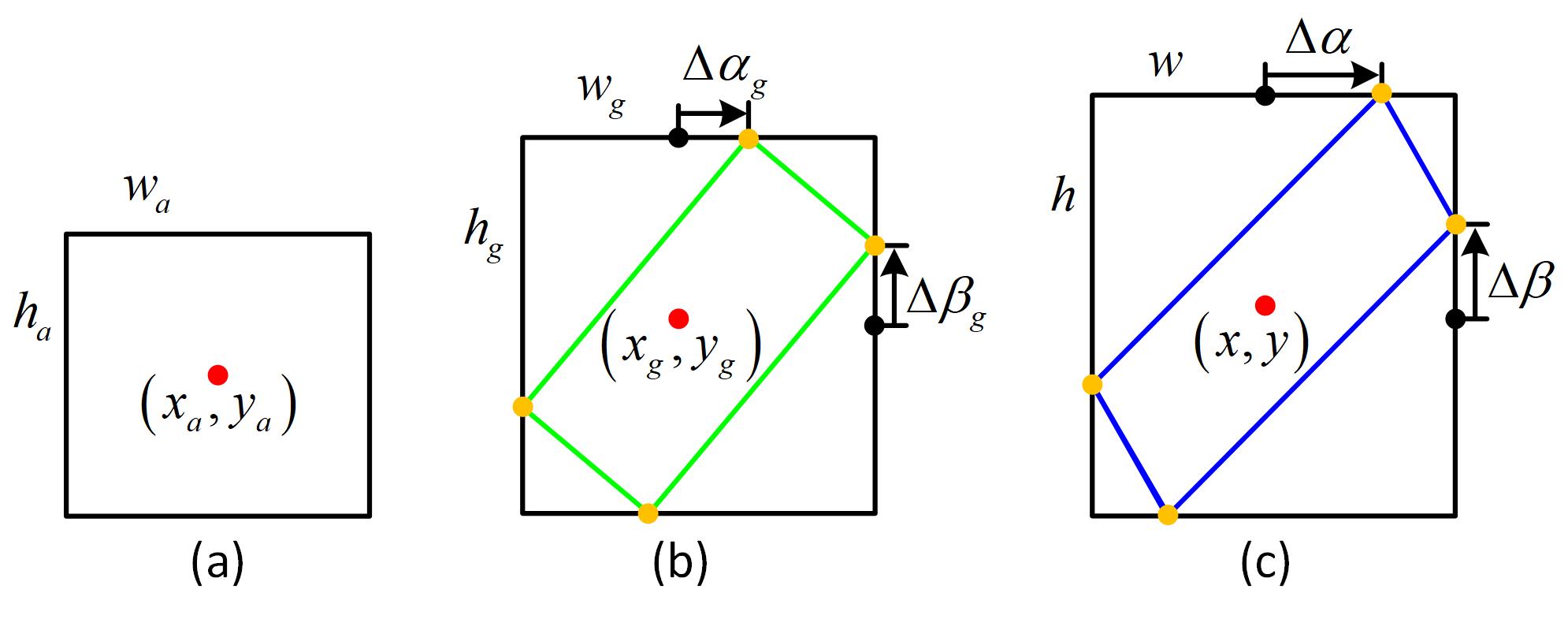}\\
	\end{center}
	\vspace*{-5mm}
	\caption{The illustration of box-regression parameterization. Black dots are the midpoints of the top and right sides, and orange dots are the vertexes of the oriented bounding box. (a) Anchor. (b) Ground-truth box. (c) Predicted box.}\label{Fig4}
	\vspace*{-3mm}
\end{figure}

Next, we define the loss function $L_{1}$ as follows:
\begin{equation}
L_{1}=\frac{1}{N} \sum_{i=1}^{N} F_{c l s}\left(p_{i}, p_{i}^{*}\right)+\frac{1}{N} p_{i}^{*} \sum_{i=1}^{N} F_{r e g}\left(\boldsymbol{\delta}_{i}, \boldsymbol{t}_{i}^{*}\right)
\end{equation}
Here, $i$ is the index of the anchors and $N$ (by default $N$=256) is the total number of samples in a mini-batch. $p_{i}^{*}$ is the ground-truth label of the $i$-th anchor. $p_{i}$ is the output of the classification branch of oriented RPN, which denotes the probability that the proposal belongs to the foreground. $\boldsymbol{t}_{i}^{*}$ is the supervision offset of the ground-truth box relative to $i$-th anchor, which is a parameterized 6-dimensional vector $\boldsymbol{t}_{i}^{*}=\left(t_{x}^{*}, t_{y}^{*}, t_{w}^{*}, t_{h}^{*}, t_{\alpha}^{*}, t_{\beta}^{*}\right)$ from the regression branch of oriented RPN, denoting the offset of the predicted proposal relative to the $i$-th anchor. $F_{c l s}$ is the cross entropy loss. $F_{r e g}$ is the Smooth $L 1$ loss. For box regression (see Figure \ref{Fig4}), we adopt the affine transformation, which is formulated as follows:
\begin{equation}
\left\{\begin{array}{ll}
\delta_{\alpha}=\Delta \alpha / w, & \delta_{\beta}=\Delta \beta / h \\
\delta_{w}=\log \left(w / w_{a}\right), & \delta_{h}=\log \left(h / h_{a}\right) \\
\delta_{x}=\left(x-x_{a}\right) / w_{a}, & \delta_{y}=\left(y-y_{a}\right) / h_{a} \\
t_{\alpha}^{*}=\Delta \alpha_{g} / w_{g}, & t_{\beta}^{*}=\Delta \beta_{g} / h_{g} \\
t_{w}^{*}=\log \left(w_{g} / w_{a}\right), & t_{h}^{*}=\log \left(h_{g} / h_{a}\right) \\
t_{x}^{*}=\left(x_{g}-x_{a}\right) / w_{a}, & t_{y}^{*}=\left(x_{g}-x_{a}\right) / h_{a}
\end{array}\right.
\end{equation}
where $\left(x_{g}, y_{g}\right)$, $w_{g}$ and $h_{g}$ are the center coordinate, width, and height of external rectangle, respectively. $\Delta \alpha_{g}$ and $\Delta \beta_{g}$ are the offsets of the top and right vertexes relative to the midpoints of top and left sides.
\begin{figure}
	\begin{center}
		\includegraphics[width=0.90\linewidth]{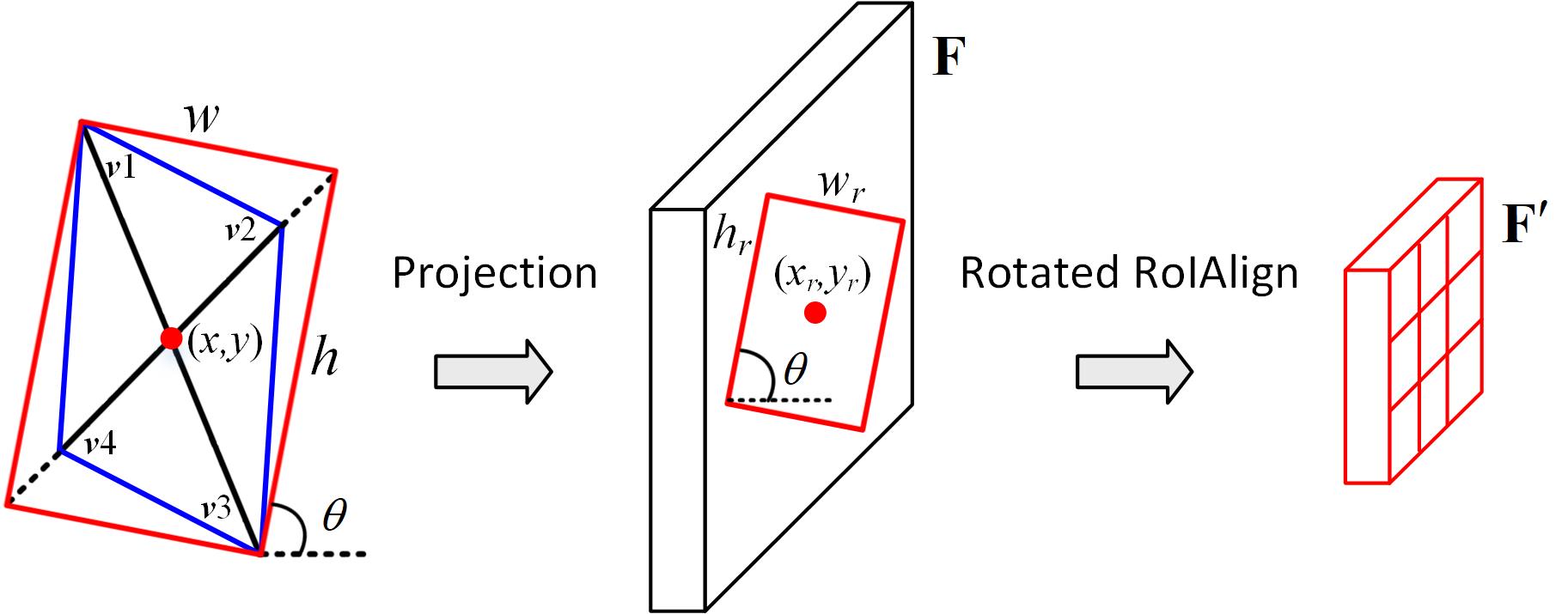}\\
	\end{center}
	\vspace*{-5mm}
	\caption{Illustration of the process of rotated RoIAlign. Blue box is a parallelogram proposal generated by oriented RPN, and the most-left red box is its corresponding rectangular proposal used for projection and rotated RoIAlign.}\label{Fig5}
	\vspace{-3mm}
\end{figure}
\subsection{Oriented R-CNN Head}
Oriented R-CNN head takes the feature maps $\left\{P_{2}, P_{3}, P_{4}, P_{5}\right\}$ and a set of oriented proposals as input. For each oriented proposal, we use rotated RoI alignment (rotated RoIAlign for short) to extract a fixed-size feature vector from its corresponding feature map. Each feature vector is fed into two fully-connected layers (FC1 and FC2, see Figure \ref{Fig2}), followed by two sibling fully-connected layers: one outputs the probability that the proposal belongs to \textit{K}+1 classes (\textit{K} object classes plus 1 background class) and the other one produces the offsets of the proposal for each of the \textit{K} object classes.
\vspace*{-3mm}
\subsubsection{Rotated RoIAlign}
Rotated RoIAlign is an operation for extracting rotation-invariant features from each oriented proposal. Now, we describe the process of rotated RoIAlign according to Figure 5. The oriented proposal generated by oriented RPN is usually a parallelogram (blue box in Figure \ref{Fig5}), which is denoted with the parameters $\mathbf{v}=(\boldsymbol{v} 1, \boldsymbol{v} 2, \boldsymbol{v} 3, \boldsymbol{v} 4)$, where $\boldsymbol{v}1$, $\boldsymbol{v}2$, $\boldsymbol{v}3$, and $\boldsymbol{v}4$ are its vertex coordinates. For ease of computing, we need to adjust each parallelogram to a rectangular with direction. To be specific, we achieve this by extending the shorter diagonal (the line from $\boldsymbol{v}2$ to $\boldsymbol{v}4$ in Figure 5) of the parallelogram to have the same length as the longer diagonal. After this simple operation, we obtain the oriented rectangular $(x, y, w, h, \theta)$ (red box in Figure 5) from the parallelogram, where $\theta \in[-\pi / 2, \pi / 2]$ is defined by the intersection angle between the horizontal axis and the longer side of the rectangular.

We next project the oriented rectangular $(x, y, w, h, \theta)$ to the feature map $\mathbf{F}$ with the stride of $s$ to obtain a rotated RoI, which is defined by $\left(x_{r}, y_{r}, w_{r}, h_{r}, \theta\right)$ through the following operation.
\begin{equation}
\left\{\begin{array}{l}
w_{r}=w / s, \quad h_{r}=h / s \\
x_{r}=\lfloor x / s\rfloor, \quad y_{r}=\lfloor y / s\rfloor
\end{array}\right.
\end{equation}

Then, each rotated RoI is divided into $m$$\times$$m$ grids ($m$ defaults to 7) to get a fixed-size feature map $\mathbf{F}^{\prime}$ with the dimension of $m$$\times$$m$$\times$$C$. For each grid with index $(i, j)$ $(0 \leq i, j \leq m-1)$     in the $c$-th channel $(1 \leq c<C)$, its value is calculated as follows:
\begin{equation}
\mathbf{F}_{c}^{\prime}(i, j)=\sum_{(x, y) \in \operatorname{area}(i, j)} \mathbf{F}_{c}(R(x, y, \theta)) / n
\end{equation}
where $\mathbf{F}_{c}$ is the feature of the $c$-th channel, $n$ is the number of samples localized within each grid, and $\operatorname{area}(i, j)$ is the coordinate set contained in the grid with index $(i, j)$. $R(\cdot)$ is a rotation transformation the same as~\cite{ding2019}.

\begin{figure*}
	\begin{center}
		\includegraphics[width=0.90\linewidth]{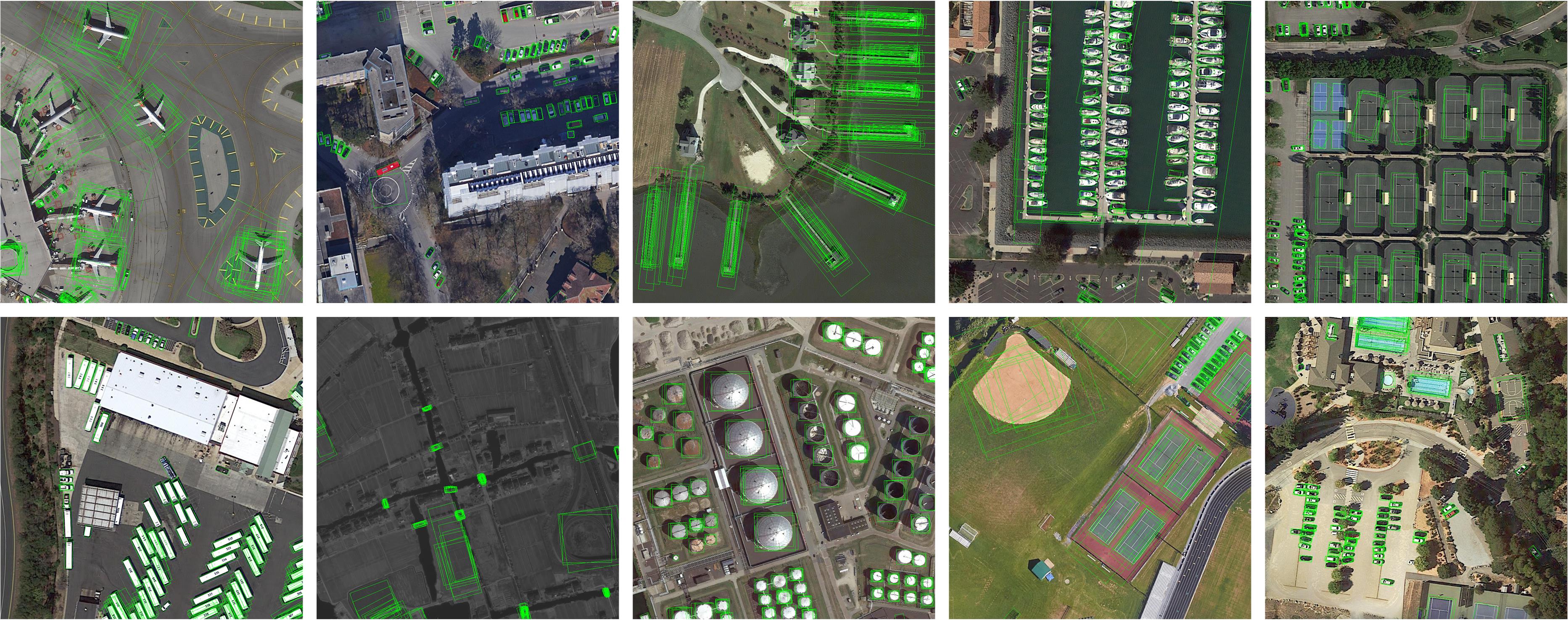}\\
	\end{center}
    \vspace*{-5.1mm}
	\caption{Proposals generated by oriented RPN on the DOTA dataset. The top-200 proposals per image are displayed.}\label{Fig6}
	\vspace{-0.5mm}
\end{figure*}

\subsection{Implementation Details}
Oriented R-CNN is trained in an end-to-end manner by jointly optimizing oriented RPN and oriented R-CNN head. During inference, the oriented proposals generated by oriented RPN generally have high overlaps. In order to reduce the redundancy, we remain 2000 proposals per FPN level in the first stage, followed by Non-Maximum Suppression (NMS). Considering the inference speed, the horizontal NMS with the IoU threshold of 0.8 is adopted. We merge the remaining proposals from all levels, and choose top-1000 ones based on their classification scores as the input of the second stage. In the second stage, ploy NMS for each object class is performed on those predicted oriented bounding boxes whose class probability is higher than 0.05. The ploy NMS IoU threshold is 0.1.

\begin{figure*}
	\begin{center}
		\includegraphics[width=0.95\linewidth]{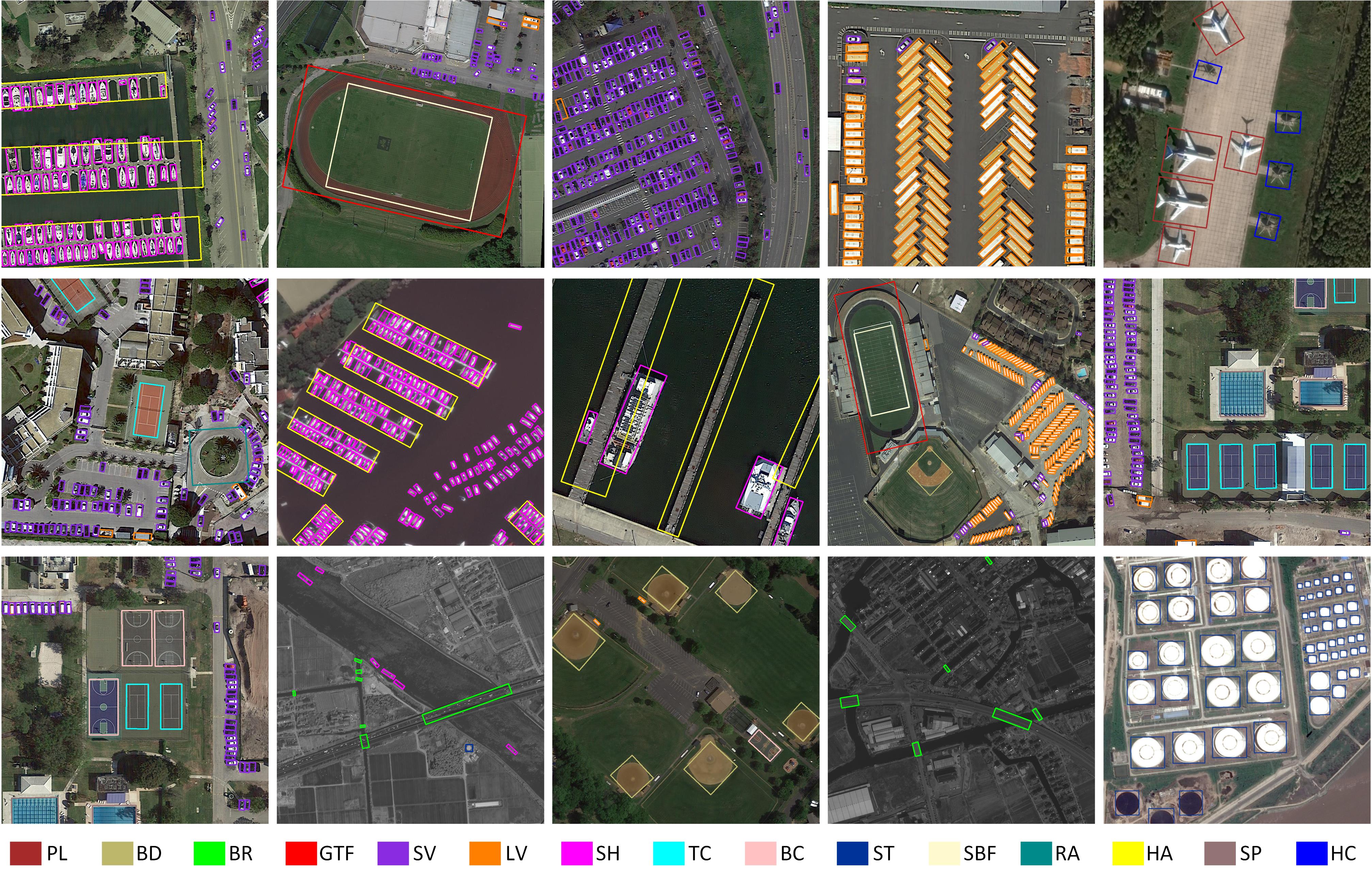}\\
	\end{center}
    \vspace*{-5mm}
	\caption{Examples of detection results on the DOTA dataset using oriented R-CNN with R-50-FPN backbone. The confidence threshold is set to 0.3 when visualizing these results. One color stands for one object class.}\label{Fig7}
	\vspace{-3mm}
\end{figure*}

\begin{figure*}[t]
	\begin{center}
		\includegraphics[width=0.95\linewidth]{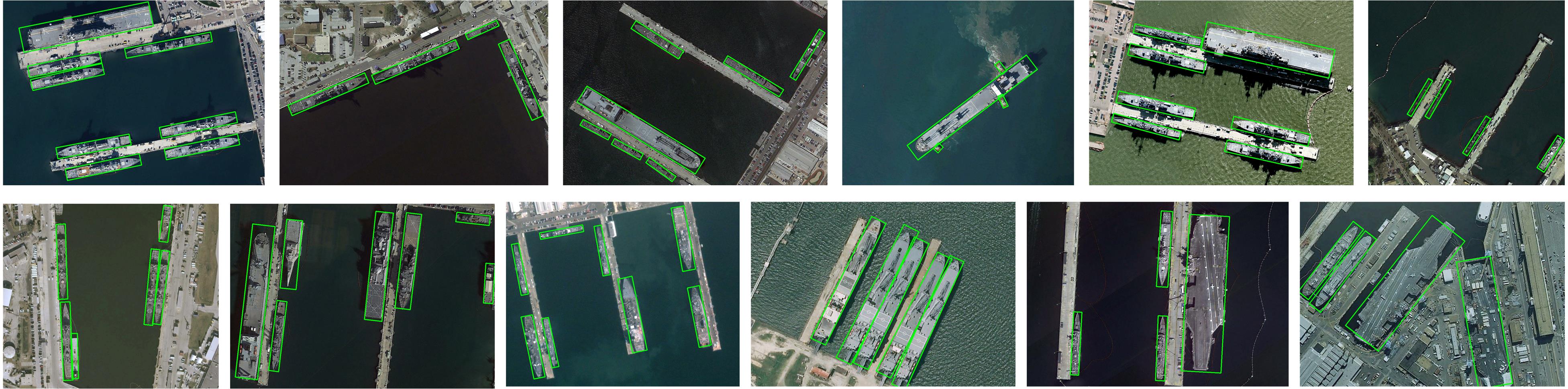}\\
	\end{center}
    \vspace*{-5mm}
	\caption{Examples of detection results on the HRSC2016 dataset using oriented R-CNN with R-50-FPN backbone. The oriented bounding boxes whose scores are higher than 0.3 are shown.}\label{Fig8}
	\vspace{0.5mm}
\end{figure*}

\section{Experiments}
To evaluate our proposed method, we conduct extensive experiments on two most widely-used oriented object detection datasets, namely DOTA~\cite{xia2018dota} and HRSC2016~\cite{liu2016ship}.

\subsection{Datasets}
DOTA is a large-scale dataset for oriented object detection. It contains 2806 images and 188282 instances with oriented bounding box annotations, covered by the following 15 object classes: Bridge (BR), Harbor (HA), Ship (SH), Plane (PL), Helicopter (HC), Small vehicle (SV), Large vehicle (LV), Baseball diamond (BD), Ground track field (GTF), Tennis court (TC), Basketball court (BC), Soccer-ball field (SBF), Roundabout (RA), Swimming pool (SP), and Storage tank (ST). The image size of the DOTA dataset is large: from 800$\times$800 to 4000$\times$4000 pixels. We use training and validation sets for training and the rest for testing. The detection accuracy is obtained by submitting testing results to DOTA's evaluation server.

HRSC2016 is another widely-used dataset for arbitrary-oriented ship detection. It contains 1061 images with the size ranging from 300$\times$300 to 1500$\times$900. Both the training set (436 images) and validation set (181 images) are used for training and the remaining for testing. For the detection accuracy on the HRSC2016, we adopt the mean average precision (mAP) as evaluation criteria, which is consistent with PASCAL VOC 2007~\cite{everingham2010pascal} and VOC 2012.
\begin{table}[h]\small
	\begin{center}
		\renewcommand\arraystretch{1.3}
		\resizebox{0.4\textwidth}{!}{
			\begin{tabular}{|p{2cm}<{\centering}|p{2cm}<{\centering}|p{1.5cm}<{\centering}|p{1.5cm}<{\centering}|}
				\hline
				Method       & $\mathrm{R}_{300}$ & $\mathrm{R}_{1000}$ & $\mathrm{R}_{2000}$ \\ \hline
				\hline
				Oriented RPN & 81.60              & 92.20               & \textbf{92.80}               \\ \hline
		\end{tabular}}
	\end{center}
	\vspace*{-5mm}
	\caption{Recall results on the DOTA validation set.}\label{table1}
	\vspace{-6mm}
\end{table}
\begin{table*}[]\small
	\begin{center}
		\renewcommand\arraystretch{1.3}
		\resizebox{0.95\textwidth}{!}{
			\begin{tabular}{|p{2.8cm}<{\centering}|p{1.5cm}<{\centering}|p{0.8cm}<{\centering}|p{0.8cm}<{\centering}|p{0.8cm}<{\centering}|p{0.8cm}<{\centering}|p{0.8cm}<{\centering}|p{0.8cm}<{\centering}|p{0.8cm}<{\centering}|p{0.8cm}<{\centering}|p{0.8cm}<{\centering}|p{0.8cm}<{\centering}|p{0.8cm}<{\centering}|p{0.8cm}<{\centering}|p{0.8cm}<{\centering}|p{0.8cm}<{\centering}|p{0.8cm}<{\centering}|p{0.8cm}<{\centering}|}
				\hline
				Method                                 & Backbone  & PL             & BD             & BR             & GTF            & SV             & LV             & SH             & TC             & BC             & ST             & SBF            & RA             & HA             & SP             & HC             & mAP            \\ \hline
				\textbf{One-stage}                     & \multicolumn{17}{c|}{}                                                                                                                                                                                                                                                                    \\ \hline
				RetinaNet-O$^\dag$                   & R-50-FPN  & 88.67          & 77.62          & 41.81          & 58.17          & 74.58          & 71.64          & 79.11          & 90.29          & 82.18          & 74.32          & 54.75          & 60.60          & 62.57          & 69.67          & 60.64          & 68.43          \\ 
				DRN~\cite{pan2020dynamic}              & H-104     & 88.91          & 80.22          & 43.52          & 63.35          & 73.48          & 70.69          & 84.94          & 90.14          & 83.85          & 84.11          & 50.12          & 58.41          & 67.62          & 68.60          & 52.50          & 70.70          \\ 
				R3Det~\cite{yang2019r3det}             & R-101-FPN & 88.76          & 83.09          & 50.91          & 67.27          & 76.23          & 80.39          & 86.72          & 90.78          & 84.68          & 83.24          & 61.98          & 61.35          & 66.91          & 70.63          & 53.94          & 73.79          \\ 
				PIoU~\cite{chen2020piou}               & DLA-34    & 80.90          & 69.70          & 24.10          & 60.20          & 38.30          & 64.40          & 64.80          & 90.90          & 77.20          & 70.40          & 46.50          & 37.10          & 57.10          & 61.90          & 64.00          & 60.50          \\ 
				RSDet~\cite{qian2019learning}          & R-101-FPN & 89.80          & 82.90          & 48.60          & 65.20          & 69.50          & 70.10          & 70.20          & 90.50          & 85.60          & 83.40          & 62.50          & 63.90          & 65.60          & 67.20          & 68.00          & 72.20          \\ 
				DAL~\cite{ming2020dynamic}             & R-50-FPN  & 88.68          & 76.55          & 45.08          & 66.80          & 67.00          & 76.76          & 79.74          & 90.84          & 79.54          & 78.45          & 57.71          & 62.27          & 69.05          & 73.14          & 60.11          & 71.44          \\ 
				S$^{2}$ANet~\cite{han2021}             & R-50-FPN  & 89.11          & 82.84          & 48.37          & 71.11          & 78.11          & 78.39          & 87.25          & 90.83          & 84.90          & 85.64          & 60.36          & 62.60          & 65.26          & 69.13          & 57.94          & 74.12          \\ \hline
				\textit{\textbf{two-stage}}            & \multicolumn{17}{c|}{}                                                                                                                                                                                                                                                                    \\ \hline
				ICN~\cite{azimi2018towards}            & R-101-FPN & 81.36          & 74.30          & 47.70          & 70.32          & 64.89          & 67.82          & 69.98          & 90.76          & 79.06          & 78.20          & 53.64          & 62.90          & 67.02          & 64.17          & 50.23          & 68.16          \\ 
				Faster R-CNN-O$^\dag$                & R-50-FPN  & 88.44          & 73.06          & 44.86          & 59.09          & 73.25          & 71.49          & 77.11          & 90.84          & 78.94          & 83.90          & 48.59          & 62.95          & 62.18          & 64.91          & 56.18          & 69.05          \\ 
				CAD-Net~\cite{zhang2019cad}            & R-101-FPN & 87.80          & 82.40          & 49.40          & 73.50          & 71.10          & 63.50          & 76.60          & 90.90          & 79.20          & 73.30          & 48.40          & 60.90          & 62.00          & 67.00          & 62.20          & 69.90          \\ 
				RoI Transformer~\cite{ding2019}        & R-101-FPN & 88.64          & 78.52          & 43.44          & 75.92          & 68.81          & 73.68          & 83.59          & 90.74          & 77.27          & 81.46          & 58.39          & 53.54          & 62.83          & 58.93          & 47.67          & 69.56          \\ 
				SCRDet~\cite{yang2019}                                 & R-101-FPN & 89.98          & 80.65          & 52.09          & 68.36          & 68.36          & 60.32          & 72.41          & 90.85          & \textbf{87.94} & 86.86          & 65.02          & 66.68          & 66.25          & 68.24          & 65.21          & 72.61          \\ 
				RoI Transformer$^\textbf{+}$                       & R-50-FPN  & 88.65          & 82.60          & 52.53          & 70.87          & 77.93          & 76.67          & 86.87          & 90.71          & 83.83          & 82.51          & 53.95          & 67.61          & 74.67          & 68.75          & 61.03          & 74.61          \\ 
				Gliding Vertex~\cite{xu2020}           & R-101-FPN & 89.64          & 85.00          & 52.26          & 77.34          & 73.01          & 73.14          & 86.82          & 90.74          & 79.02          & 86.81          & 59.55          & 70.91          & 72.94          & 70.86          & 57.32          & 75.02          \\ 
				FAOD~\cite{li2019feature}              & R-101-FPN & 90.21          & 79.58          & 45.49          & 76.41          & 73.18          & 68.27          & 79.56          & 90.83          & 83.40          & 84.68          & 53.40          & 65.42          & 74.17          & 69.69          & 64.86          & 73.28          \\ 
				CenterMap-Net~\cite{wang2020learning}  & R-50-FPN  & 88.88          & 81.24          & 53.15          & 60.65          & 78.62          & 66.55          & 78.10          & 88.83          & 77.80          & 83.61          & 49.36          & 66.19          & 72.10          & 72.36          & 58.70          & 71.74          \\ 
				FR-Est~\cite{fu2020point}              & R-101-FPN & 89.63          & 81.17          & 50.44          & 70.19          & 73.52          & 77.98          & 86.44          & 90.82          & 84.13          & 83.56          & 60.64          & 66.59          & 70.59          & 66.72          & 60.55          & 74.20          \\ 
				Mask OBB~\cite{wang2019mask}           & R-50-FPN  & 89.61          & 85.09          & 51.85          & 72.90          & 75.28          & 73.23          & 85.57          & 90.37          & 82.08          & 85.05          & 55.73          & 68.39          & 71.61          & 69.87          & 66.33          & 74.86          \\ \hline
				\textit{\textbf{Ours}}                 & \multicolumn{17}{c|}{}                                                                                                                                                                                                                                                                    \\ \hline
				Oriented R-CNN                         & R-50-FPN  & 89.46          & 82.12          & 54.78          & 70.86          & 78.93          & 83.00          & 88.20          & 90.90          & 87.50          & 84.68          & 63.97          & 67.69          & 74.94          & 68.84          & 52.28          & 75.87          \\ 
				Oriented R-CNN                         & R-101-FPN & 88.86          & 83.48          & 55.27          & 76.92          & 74.27          & 82.10          & 87.52          & \textbf{90.90} & 85.56          & 85.33          & 65.51          & 66.82          & 74.36          & 70.15          & 57.28          & 76.28          \\ 
				Oriented R-CNN$^\ddag$                 & R-50-FPN  & 89.84          & \textbf{85.43} & 61.09          & 79.82          & \textbf{79.71} & \textbf{85.35} & \textbf{88.82} & 90.88          & 86.68          & 87.73          & 72.21          & \textbf{70.80} & 82.42          & 78.18          & \textbf{74.11} & \textbf{80.87} \\ 
				Oriented R-CNN$^\ddag$          & R-101-FPN & \textbf{90.26} & 84.74          & \textbf{62.01} & \textbf{80.42} & 79.04          & 85.07          & 88.52          & 90.85          & 87.24          & \textbf{87.96} & \textbf{72.26} & 70.03          & \textbf{82.93} & \textbf{78.46} & 68.05          & 80.52          \\ \hline
		\end{tabular}}
	\end{center}
	\vspace*{-4mm}
	\caption{Comparison with state-of-the-art methods on the DOTA dataset. $\dag$ means the results from AerialDetection (the same below). $\ddag$ denotes multi-scale training and testing.}\label{table2}
	\vspace{-3mm}
\end{table*}
\subsection{Parameter settings}
\vspace{-1mm}
We use a single RTX 2080Ti with the batch size of 2 for training. The inference time is also tested on a single RTX 2080Ti. The experimental results are produced on the mmdetection platform~\cite{chen2019mmdetection}. ResNet50~\cite{he2016identity} and ResNet101~\cite{he2016identity} are used as our backbones. They are pre-trained on ImageNet~\cite{deng2009imagenet}. Horizontal and vertical flipping are adopted as data augmentation during training. We optimize the overall network with SGD algorithm with the momentum of 0.9 and the weight decay of 0.0001.
On the DOTA dataset, we crop the original images into 1024$\times$1024 patches. The stride of cropping is set to 824, that is, the pixel overlap between two adjacent patches is 200. With regard to multi-scale training and testing, we first resize the original images at three scales (0.5, 1.0 and 1.5) and crop them into 1024$\times$1024 patches with the stride of 524. We train oriented R-CNN with 12 epochs. The initial learning rate is set to 0.005 and divided by 10 at epoch 8 and 11. The ploy NMS threshold is set to 0.1 when merging image patches.

For the HRSC2016 dataset, we do not change the aspect ratios of images. The shorter sides of the images are resized to 800 while the longer sides are less than or equal to 1333. During training, 36 epochs are adopted. The initial learning rate is set to 0.005 and divided by 10 at epoch 24 and 33.

\subsection{Evaluation of Oriented RPN}
We evaluate the performance of oriented RPN in terms of recall. The results of oriented RPN are reported on the DOTA validation set, and ResNet-50-FPN is used as the backbone. To simplify the process, we just calculate the recall based on the patches cropped from original images, without merging them. The IoU threshold with ground-truth boxes is set to 0.5. We respectively select top-300, top-1000, and top-2000 proposals from each image patch to report their recall values, denoted as $\mathrm{R}_{300}$, $\mathrm{R}_{1000}$, and $\mathrm{R}_{2000}$. The results are presented in Table \ref{table1}.
As can be seen, our oriented RPN can achieve the recall of 92.80\% when using 2000 proposals. The recall drops very slightly (0.6\%) when the number of proposals changes from 2000 to 1000, but it goes down sharply when using 300 proposals. Therefore, in order to trade-off the inference speed and detection accuracy, we choose 1000 proposals as the input of oriented R-CNN head at test-time for both of the two datasets. In Figure \ref{Fig6}, we show some examples of proposals generated by oriented RPN on the DOTA dataset. The top-200 proposals per image are displayed. As shown, our proposed oriented RPN could well localize the objects no matter their sizes, aspect ratios, directions, and denseness.
\begin{table}[]\small
	\begin{center}
		\renewcommand\arraystretch{1.3}	
		\resizebox{0.42\textwidth}{!}{
			\begin{tabular}{|p{3.5cm}<{\centering}|p{1.8cm}<{\centering}|p{1.2cm}<{\centering}|p{1.2cm}<{\centering}|}
				\hline
				Method          & Backbone  & mAP(07) & mAP(12) \\ \hline
				\hline
				PIoU~\cite{chen2020piou}            & DLA-34    & 89.20   & -       \\ 
				DRN~\cite{pan2020dynamic}             & H-34      & -       & 92.70   \\ 
				R3Det~\cite{yang2019r3det}           & R-101-FPN & 89.26   & 96.01   \\ 
				DAL~\cite{ming2020dynamic}             & R-101-FPN & 89.77   & -       \\ 
				S$^{2}$ANet~\cite{han2021}     & R-101-FPN & 90.17   & 95.01   \\ 
				Rotated RPN~\cite{ma2018}     & R-101     & 79.08   & 85.64   \\ 
				R2CNN~\cite{jiang2017r2cnn}           & R-101     & 73.07   & 79.73   \\ 
				RoI Transformer~\cite{ding2019} & R-101-FPN & 86.20   & -       \\ 
				Gliding Vertex~\cite{xu2020}  & R-101-FPN & 88.20   & -       \\ 
				CenterMap-Net~\cite{wang2020learning}   & R-50-FPN  & -       & 92.80   \\ 
				Oriented R-CNN  & R-50-FPN  & 90.40   & 96.50   \\ 
				Oriented R-CNN  & R-101-FPN & \textbf{90.50}   & \textbf{97.60 }  \\ \hline
		\end{tabular}}
	\end{center}
	\vspace*{-4mm}
	\caption{Comparison results on the HRSC2016 dataset.}\label{table3}
	\vspace{-0.5mm}
\end{table}
\begin{table}[]\small
	\begin{center}
		\renewcommand\arraystretch{1.38}	
		\resizebox{0.4\textwidth}{!}{
			\begin{tabular}{|p{3cm}<{\centering}|p{2cm}<{\centering}|p{1.5cm}<{\centering}|p{1.5cm}<{\centering}|p{3cm}<{\centering}}
				\hline
				Method           & Framwork  & FPS  & mAP   \\ \hline
				\hline
				RetinaNet-O$^\dag$      & One-stage & \textbf{16.1} & 68.43 \\ 
				S$^{2}$ANet~\cite{han2021}      & One-stage & 15.3 & 74.12 \\ 
				Faster R-CNN-O$^\dag$   & Two-stage & 14.9 & 69.05 \\ 
				RoI Transformer$^\textbf{+}$ & Two-stage & 11.3 & 74.61 \\ 
				Oriented R-CNN   & Two-stage & 15.1 & \textbf{75.87} \\ \hline
		\end{tabular}}
	\end{center}
	\vspace*{-3.7mm}
	\caption{Speed versus accuracy on the DOTA dataset.}\label{table4}
	\vspace{-3mm}
\end{table}
\subsection{Comparison with State-of-the-Arts}
We compare our oriented R-CNN method with 19 oriented object detection methods for the DOTA dataset, and 10 methods for the HRSC2016 dataset. Table \ref{table2} and Table \ref{table3} report the detailed comparison results on the DOTA  and  HRSC2016 datasets, respectively. The backbones are as follows: R-50 stands for ResNet-50, R-101 denotes ResNet-101, H-104 is the 104-layer hourglass network~\cite{yang2017stacked}, and DLA-34 means the 34-layer deep layer aggregation network~\cite{zhou2019objects}.

On the DOTA dataset, our method surpasses all comparison methods. With R-50-FPN and R-101-FPN as the backbones, our method obtains 75.87\% and 76.28\% mAP, respectively. It is surprising that using the backbone of R-50-FPN, we even outperform all comparison methods with the R-101-FPN backbone. In addition, with multi-scale training and testing strategies, our method reaches 80.87\% mAP using R-50-FPN backbone, which is very competitive compared to the current state-of-the-art methods. Figure \ref{Fig7} shows some results on the DOTA dataset.

\begin{figure}[th]
	\begin{center}
		\includegraphics[width=0.78\linewidth]{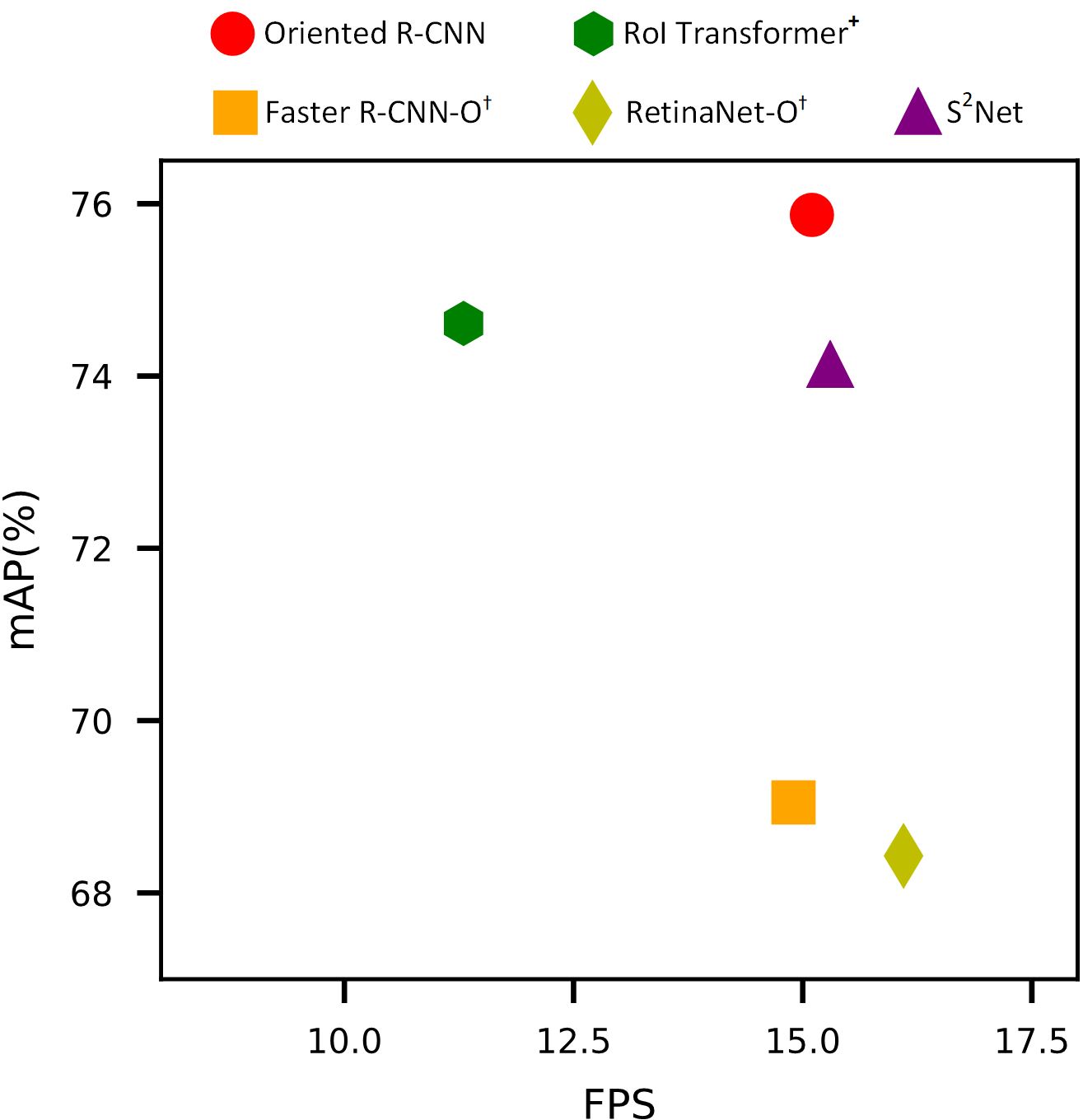}\\
	\end{center}
	\vspace*{-5mm}
	\caption{Speed versus accuracy on the DOTA test set.}\label{Fig9}
	\vspace{-3mm}
\end{figure}
For the HRSC2016 dataset, we list the mAP values of different methods under PASCAL VOC 2007 and VOC 2012 metrics. With R-50-FPN and R-101-FPN, our oriented R-CNN both achieves the best accuracy. Some visualization results are presented in Figure \ref{Fig8}.

\subsection{Speed versus Accuracy}
Under the same setting, we compare the speed and accuracy of different methods. The comparison results are presented in Table 4. All methods adopt R-50-FPN as the backbone. The hardware platform of testing is a single RTX 2080Ti with batch size of 1. During testing, the size of input images is 1024$\times$1024. As shown in Table 4, our method has higher detection accuracy (75.87\% mAP) than other methods but runs with comparable speed (15.1 FPS). The speed of oriented R-CNN is almost close to one-stage detectors, but the accuracy is much higher than one-stage detectors (see Table \ref{table4} and Figure \ref{Fig9}).

\section{Conclusions}
This paper proposed a practical two-stage detector, named oriented R-CNN, for arbitrary-oriented object detection in images. We conduct extensive experiments on two challenging oriented object detection benchmarks. Experimental results show that our method has competitive accuracy to the current advanced two-stage detectors, while keeping comparable efficiency compared with one-stage oriented detectors.

\section*{Acknowledgments}
This work was supported by the National Natural Science Foundation of China (No. 61772425), the Shaanxi Science Foundation for Distinguished Young Scholars (No. 2021JC-16), and the Doctorate Foundation of Northwestern Polytechnical University (No. CX2021082).
{\small
\bibliographystyle{ieee_fullname}
\bibliography{egbib}
}

\end{document}